%% file: report.tex
\documentclass[
10pt, % Main document font size
a4paper, % Paper type, use 'letterpaper' for US Letter paper
oneside, % One page layout (no page indentation)
headinclude,footinclude, % Extra spacing for the header and footer
BCOR5mm, % Binding correction
]{scrartcl}

\input{structure.tex} % Include the structure.tex file which specified the document structure and layout

\title{\normalfont\spacedallcaps{Modern Methods of Text Generation}} % The article title

\subtitle{National Research University Higher School of Economics\\Moscow, Russia} % Uncomment to display a subtitle

\author{\spacedlowsmallcaps{Dimas Munoz Montesinos}}

\date{Aug 31th, 2020} % An optional date to appear under the author(s)

\begin{document}
	
	%----------------------------------------------------------------------------------------
	%	HEADERS
	%----------------------------------------------------------------------------------------
	
	\renewcommand{\sectionmark}[1]{\markright{\spacedlowsmallcaps{#1}}} % The header for all pages (oneside) or for even pages (twoside)
	\lehead{\mbox{\llap{\small\thepage\kern1em\color{halfgray} \vline}\color{halfgray}\hspace{0.5em}\rightmark\hfil}} % The header style
	
	\pagestyle{scrheadings} % Enable the headers specified in this block
	
	%----------------------------------------------------------------------------------------
	%	COVER
	%----------------------------------------------------------------------------------------
	
	\maketitle % Print the title/author/date block
	
	%----------------------------------------------------------------------------------------
	%	ABSTRACT, TABLE OF CONTENTS & LISTS OF FIGURES AND TABLES
	%----------------------------------------------------------------------------------------
	
	\section*{Abstract}
	
	Synthetic text generation is challenging and has limited success. Recently, a new architecture, called Transformers, allow machine learning models to understand better sequential data, such as translation or summarization. BERT and GPT-2, using Transformers in their cores, have shown a great performance in tasks such as text classification, translation and NLI tasks. In this article, we analyse both algorithms and compare their output quality in text generation tasks.

	\setcounter{tocdepth}{2} % Set the depth of the table of contents to show sections and subsections only
	
	\tableofcontents % Print the table of contents
	
	%----------------------------------------------------------------------------------------
	
	\newpage % Start the article content on the second page, remove this if you have a longer abstract that goes onto the second page
	
	%----------------------------------------------------------------------------------------
	%	INTRODUCTION
	%----------------------------------------------------------------------------------------
	
	\section{Introduction}
	
	Natural Language Processing (NLP) is a large field where we can find different tasks: Text Classification, Named Entities Recognition, Language Translation... These tasks have a common challenge: texts written with human languages (usually unstructured texts). The task that concerns us in this article is text generation using a Conditional Language Model and the novel Transformers architecture.
	
	In order to understand Text Generation, it is necesary to define what is a Language Model (LM). From Wikipedia, ``\textit{a Statistical Language Model is a probability distribution over sequences of words, such that, given a sequence of length m, it assigns a probability $P(w_{1}, \ldots, w_{m})$ to the whole sequence.}''. In consequence, we can use a Conditional LM to find the probability of the next word in a sequence: $P(w_{m+1} | w_{1}, \ldots, w_{m})$.
	
	In this article, we assume that you have fundamental knowledge of deep learning, word vectors and embedding space. Nevertheless, here we describe some models and techniques which are relevant to understand Transformer-based models.
	
	\subsubsection*{Seq2seq models}
	
	During a long time, Conditional Text Generation was based in Seq2Seq models \cite{sutskever2014sequence}. The idea behind Seq2Seq consists of 2 Recurrent Neural Networks (RRN) that try to predict the next state sequence from the previous one (the two RNNs receive the names encoder and decoder respectively). Two of the most extended RNNs were LSTM, introduced in 1997 by Sepp Hochreiter and Jurgen Schmidhuber \cite{doi:10.1162/neco.1997.9.8.1735}, and GRU, introduced in 2014 by Junyoung Chung et al \cite{chung2014empirical}. However, texts generated with RNNs are far from being perfect: they tend to be nonsense and sometimes they include spelling mistakes. Basically, one wrong prediction has the potential to make the entire sentence meaningless. Furthermore, it is not possible to apply parallelization since the RNNs need to process data as a sequence.
	
	\begin{figure}[ht]
		\centering
		\includegraphics[width=1.0\textwidth]{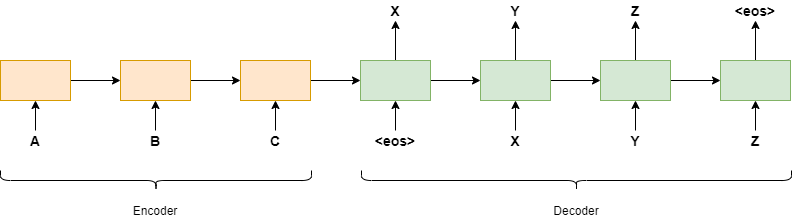}
		\caption{This is the architecture of a Seq2seq model: the first RNN is called encoder and the second one is decoder. In this case, the model receives an input sentence ``ABC'' and produces ``XYZ'' as the output sentence (the input and output may have different lengths).}
		\label{fig:Seq2seq}
	\end{figure}
	
	\subsubsection*{Contextualized word embeddings}
	
	Traditionally, a model received a sequence of tokens (usually words) and transformed them into static vectors. In short, they are simply vectors of numbers that represent the meaning of a word. One widely extended model is Word2vec (introduced by Mikolov et al. in 2013) which computes the static vector of each token (the vectors are called embeddings) \cite{mikolov2013efficient}. Furthermore, vectors of Word2vec provided State-Of-The-Art (SOTA) performance in syntactic and semantic word similarities.
	
	The power of using word vectors is that they lend themselves to mathematical operators. For example, we can add and subtract vectors:
	$$king - man + woman = queen$$
	
	The recent techniques consists of incorporating context into word embeddings: replacing static vectors with contextualized word representations has led to significant improvements on virtually every NLP task. ELMo introduced this kind of word embeddings in 2018 \cite{peters2018deep}: ``\textit{vectors are learned functions of the internal states of a deep bidirectional language model (biLM)}''. This is one of the breakthroughs which will lead models to further understanding of words. For example, they may difference homonyms (e.g. \textit{rock} can be a \textit{stone} or a \textit{music genre}) instead of having the same static vector for them.
	
	\subsubsection*{Transformers}
	
\begin{figure}[ht]
		\centering
		\includegraphics[width=0.6\textwidth]{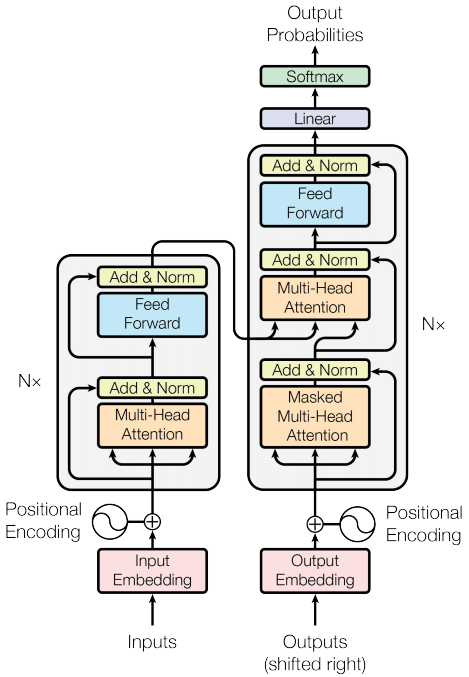}
		\caption{Transformer-model architecture described in ``\textit{Attention Is All You Need}'' \cite{vaswani2017attention}.}
		\label{fig:Transformer}
	\end{figure}

	Google's researches released a new model called Transformer in 2017 in the paper ``\textit{Attention Is All You Need}'' \cite{vaswani2017attention}. Very briefly, this architecture consists of self-attention and point-wise, fully connected layers (see Figure \ref{fig:Transformer}). Similarly to what we describe for Seq2seq, Transformers include an encoder, decoder and a final linear layer.
	
	These models are designed to handle sequences of data (especially useful in NLP). Note that, in contrast to Seq2seq they do not contain recurrence or convolution, so they do not require to process sequences in order. This fact allows us to parallelize (much more than RNNs) and reduces training time.
	
	\section{Models}
	
	\subsection{BERT}
	
	\subsubsection{Description}
	
	Bidirectional Encoder Representations from Transformers (commonly known by its abbreviated form BERT), as the name suggests, ``\textit{is designed to pretrain deep bidirectional representations from unlabeled text by jointly conditioning on both left and right context in all layers}'' \cite{devlin2018bert}. The model architecure is a technical innovation that implements multiple layers of Transformer-encoders to language modelling.
	
\begin{figure}[ht]
		\centering
		\includegraphics[width=1.0\textwidth]{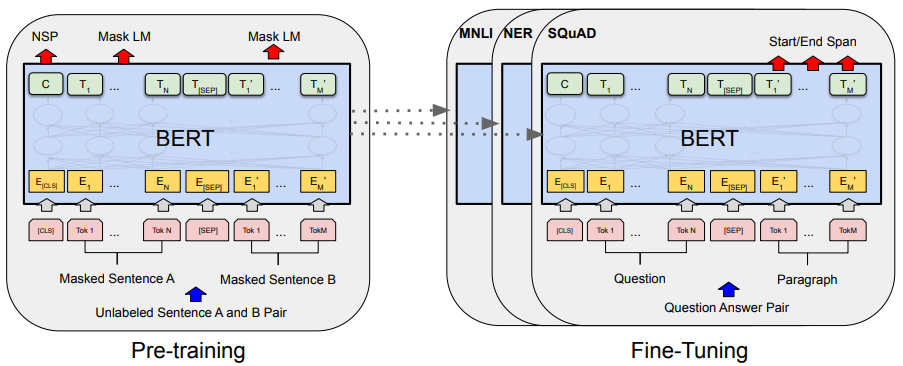}
		\caption{Overall pre-training and fine-tuning procedures described in BERT's paper \cite{devlin2018bert}.}
		\label{fig:BertSteps}
	\end{figure}

	The authors of BERT describe two steps in their framework: pre-training and fine-tuning (see Figure \ref{fig:BertSteps}).
	\begin{enumerate}
		\item During pre-training, the model is trained on unlabeled data over different pre-training tasks: Masked LM and Next Sentence Prediction. The authors pretrained BERT on the BooksCorpus ($800M$ words) and Wikipedia ($2,500M$ words).
		\item During fine-tuning, the model is initialized with the pre-trained parameters (this is also known as transfer learning). Then, each downstream task is fine-tuned separately. This process is simple, so it is not described in this article.
	\end{enumerate}
	
	\subsubsection{Input representation}
	
	In order to make BERT handle a variety of downstream tasks, the authors defined three different inputs which allow us to unambiguously represent both a single sentence and a pair of sentences.
	
	One of these inputs is the tokenized text using a technique called WordPiece tokens (i.e. sub-words units) \cite{wu2016googles}. Apart from the sub-words units, the authors introduced two new tokens that must be appended to the input sentences: $[CLS]$ in the beginning of the input and $[SEP]$ after each sentence.
	
	The second and third inputs are sequences of $0s$ and $1s$. One (called Token Type IDs or Segment IDs) indicates that a token belongs to the sentence A (a series of $0s$) or the sentence B (a series of $1s$). The other (called Mask IDs) is used when input texts are padded to same length (indicates whether the text is padded from a certain position).
	
	\begin{figure}[ht]
		\centering
		\includegraphics[width=1.0\textwidth]{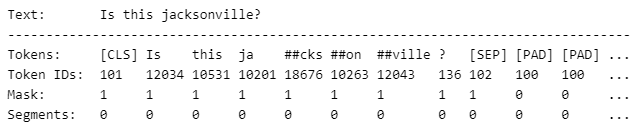}
		\caption{Example of BERT's input with a single sentence. If there were two different sentences in the input, there would be $1s$ in the \textit{segments} sequence (starting from the position of the token $[SEP]$).}
		\label{fig:BertInput}
	\end{figure}
	
	As you can see in Figure \ref{fig:BertInput}, not all tokens are words, thus it is easier to handle unknown words. Word pieces are very powerful in the sense that tokens cover all the word (even the words that do not occur in the dictionary) and we do not lose information (since all subword units are in the input).
	
	\subsubsection{Pre-training tasks}
	
	\textbf{Masked Language Model (MLM).} In this task, authors masked $15\%$ of WordPiece tokens (in each sequence of the dataset at random) and then predict those masked tokens. They only mask $15\%$ of tokens since the token $[MASK]$ is not used in fine-tuning and they may create a mismatch between pre-training and fine-tuning.
	
	\textbf{Next Sentence Prediction (NSP).} In words of the authors, ``\textit{tasks such as Question Answering (QA) and Natural Language Inference (NLI) are based on understanding the relationship between two sentences, which is not directly captured by language modeling}''. In order to train a model that understands sentence relationships, given a pair of sentences, the model should predict if the second sentence is the subsequent sentence in the original document. The authors built a dataset where $50\%$ is the actual next sentence and $50\%$ is a random sentence from a monolingual corpus.
	
	\subsection{GPT-2}
	
	\subsubsection{Description}
	
	Generative Pretrained Transformer 2, known by GPT-2, is a large unsupervised transformer-based language model and the successor to GPT \cite{radford2018gpt}. GPT-2 was introduced in June 2018 by researchers from OpenAI in their paper ``\textit{Language Models are Unsupervised Multitask Learners}'' \cite{radford2019gpt2}.
	
	GPT-2 consists of solely stacked decoder blocks from the transformer architecture. In the vanilla transformer architecture, the decoder is fed a word embedding concatenated with a context vector, both generated by the encoder. In GPT-2 the context vector is zero-initialized for the first word embedding. Furthermore, in the vanilla transformer architecture self-attention is applied to the entire surrounding context (e.g. all of the other words in the sentence), but in GPT-2 masked self-attention is used instead: the decoder is only allowed (via obfuscation masking of the remaining word positions) to glean information from the previous words in the sentence (plus the word itself). Besides this, GPT-2 is a close copy of the vanilla transformer architecture and very similar to its predecessor GPT.
	
	Authors of GPT-2 trained it with a simple objective: given some text, predict the next word. For this purpose, they used around 40GB of crawled data from internet. Also, in a similar way as BERT, they fine-tuned GPT-2 in downstream tasks to analyse the model performance in different situations.
	
	\begin{figure}[ht]
		\centering
		\includegraphics[width=1.0\textwidth]{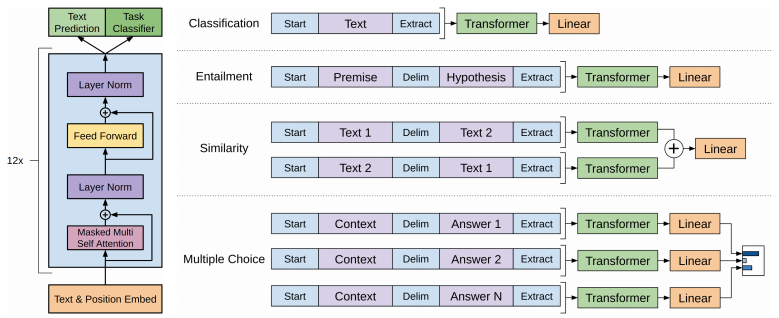}
		\caption{GPT architecture described in ``\textit{Improving Language Understanding by Generative Pre-Training}'' \cite{radford2018gpt} (transformer and training objectives are on the left, and the input transformations for fine-tuning are on the right). In GPT-2, authors moved the layer normalization to the input of each sub-block and also added another layer normalization after the final self-attention block.}
		\label{fig:GPT}
	\end{figure}
	
	\subsubsection{Input representation}
	
	GPT-2 uses Byte Pair Encoding (BPE) as input representation \cite{gage1994}. This technique allows us to combine the empirical benefits of word-level LMs with the generality of byte-level approaches. BPE is a simple compression method in which the most common pair of consecutive bytes of data is replaced with a byte that does not occur within that data. For instance, given the sequence $aaabdaaabac$ and $Z=aa$, then we obtain $ZabdZabac$.
	
	It is important to highlight the fact that the authors do not apply any kind of pre-processing to the data (e.g. lowercasing, tokenization or out-of-vocabulary tokens). They believe that ``\textit{a general language model (LM) should be able to compute the probability of (and also generate) any string}''.
	
	\section{Experiments}
	
	\subsection{Infer masked token}
	
	As we have explained before, one of the pretraining tasks of BERT was MLM so, for this experiment, we only need a pretrained BERT model and HuggingFace\footnote{\url{https://huggingface.co/models}} already provides a wide variety of pretrained and fine-tuned models in PyTorch or TensorFlow. In particular, we chose \textit{BERT multilingual base model} which is a pretrained model on the top 104 languages with the largest Wikipedia using a MLM objective. Then, we apply the softmax function to normalize the output and pick the top words.
	
	Since this experiment is really simple and does not require any further training, we proceeded to test it out.
	
	\begin{figure}[ht]
		\centering
		\includegraphics[width=0.55\textwidth]{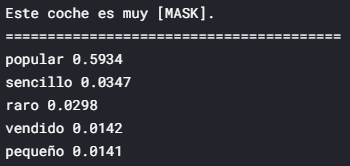}
		\caption{Given the sentence ``\textit{This car is very [MASK]}'' in Spanish, it returns words such as \textit{popular} ($59.34\%$), \textit{simple} ($3.47\%$) or \textit{uncommon} ($2.98\%$). In this case, all top-5 words make sense.}
		\label{fig:ExperimentMaskEsp}
	\end{figure}
	
	\begin{figure}[ht]
		\centering
		\includegraphics[width=0.55\textwidth]{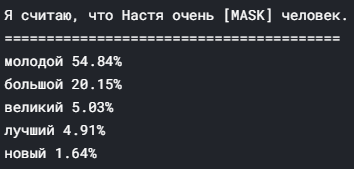}
		\caption{Given the sentence ``\textit{I think that Nastya is a very [MASK] person}'' in Russian, it returns words such as \textit{young} ($54.84\%$), \textit{big} ($20.15\%$) or \textit{great} ($5.03\%$). In this context, the last word (\textit{new}) does not make sense.}
		\label{fig:ExperimentMaskRu}
	\end{figure}
	
	Despite infered words are correct in many cases, there are other cases where the word suggestions are far to be good. For instance, the first word of ``\textit{Today is [MASK].}'' (written in English) is ``\textit{demolished}'' (it is gramatically correct, but it is very uncommon). However, if we use a only-English pretrained model (we used BERT base model), the first output is ``\textit{closed}''. Also, in non-English languages, sometimes it is unable to return a word but punctuation symbols.
	
	We believe that we must use monolingual models to obtain good results. In other words, we must train one BERT model per language. There are a wide list of arguments for this which are described in the conclusions section.
	
	You can find some infered masked tokens of our experiments in the \hyperref[sec:appendixExperiments]{appendix}. There are a wide variety of tests in three languages: English, Spanish and Russian.
	
	\subsection{Question-answering}
	
	In Question-Answering tasks, the model receives a text (called context) and a question regarding to the context, and it should mark the inital and ending of the answer in the context (n)ote that the model does not generate text).
	
	\begin{figure}[ht]
		\centering
		\includegraphics[width=0.8\textwidth]{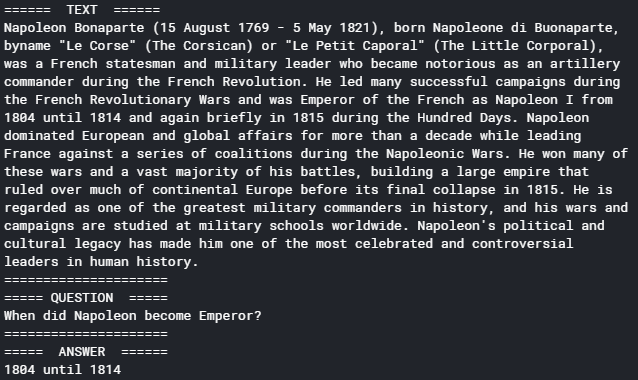}
		\caption{Sample question-answering to Napoleon's biography (context extracted from Wikipedia). The answer is good but it is not stricly accurate: it should be only ``\textit{1804}'' since we did not ask for the period.}
		\label{fig:ExperimentQaEng}
	\end{figure}
	
\begin{figure}[ht]
		\centering
		\includegraphics[width=0.65\textwidth]{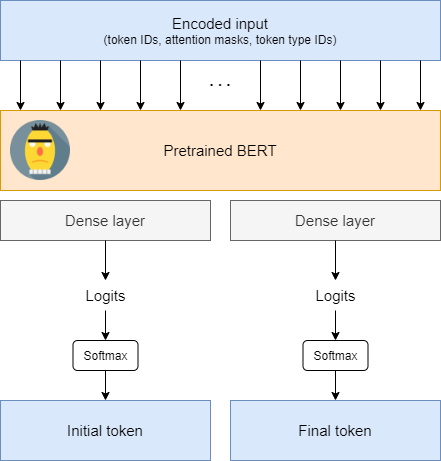}
		\caption{Architecture of QA-BERT model.}
		\label{fig:ExperimentBertQaArch}
	\end{figure}

	Similarly to the previous experiment, we import \textit{BERT multilingual base model} from HuggingFace. Then, we add two fully-connected layers to obtain the initial and ending token positions from the context (see Figure \ref{fig:ExperimentBertQaArch}). Based on these positions, we pick a portion of context and return it as the answer to the question. However, we cannot test it without previous training (fine-tuning BERT and adjust weights of fully-connected layers). In other words, we are using a technique called transfer learning: the model is pre-trained in a dataset $A$ and, then, we use that pre-trained model to carry that knowledge into solving dataset $B$.
	
	In order to train this QA model, we use XQuAD\footnote{\url{https://github.com/deepmind/xquad}}. This dataset consists of a subset of $240$ paragraphs and $1190$ question-answer pairs from the development set of SQuAD v1.1 \cite{rajpurkar2016squad} together with their professional translations into ten languages, such as Spanish and Russian.
	
	In our experiments, answers tend to be accurate when we ask simple questions and when we use similar words as the context (see Figure \ref{fig:ExperimentQaEng}). We observe that it is particularly difficult to the model to understand some synonyms and homonyms. For example, we ask (in Spanish) ``\textit{When was Napoleon crowned king of Italy?}'' and ``\textit{When did he become monarch?}'', the first gives a good answer ``\textit{March 18th, 1805}'' (strictly it was on May 26th) and the second returns a totally wrong answer ``\textit{November 11th, 1799}''.
	
	You can find some questions and answers of our experiments in the \hyperref[sec:appendixExperiments]{appendix}. There are a wide variety of questions in three languages: English, Spanish and Russian. Overall, the model answers correctly when we write questions in English.
	
	\subsection{Conditional text generation}
	
	In this task, we experimented with GPT-2 due to BERT is not designed for text generation. GPT-2 was only trained to predict the next token in a sentence, but it surprisingly learned basic competence in some tasks like translating between languages and answering questions (see Figure \ref{fig:ExperimentCTG}).
	
	\begin{figure}[ht]
		\begin{center}
			\noindent\fbox{%
				\parbox{0.98\textwidth}{%
					\textbf{The RMS Titanic was visited by divers for the first time in 14 years.} The British vessel was damaged to the bow in 1912 and the ship was found to be completely submerged. The Titanic is also the only ship that has sunk in international waters. But, according to experts, the damage to the Titanic would have been even worse, if the ship had not been in the water. In fact, according to reports, the damage could have been caused by a catastrophic failure of the [...]
				}%
			}
		\end{center}
		\caption{Text sample generated by GPT-2 after providing an initial text (in bold letters).}
		\label{fig:ExperimentCTG}
	\end{figure}
	
	GPT-2 has the ability to generate conditional text samples of unprecedented quality. In the \hyperref[sec:appendixExperiments]{appendix}, you can observe that the model is capable of generating synthetic texts close to human quality. We provided different kind of prompts to analyse the versatility of GPT-2 and, despite it shows coherence and good quality (in particular, when we write about topics highly represented in the training data), we found failures, such as sudden topic switching, repetitive text and LM failures. For example, sometimes the model negates what it just wrote in the previous sentences. In Figure \ref{fig:ExperimentCTG}, even though text quality is good, you can observe that it also wrote the extravagant sentence ``\textit{if the ship had not been in the water}''
	
	As we explain later, GPT-2 and BERT architectures are different (the first is a Transformer-encoder architecture and the second is a Transformer-decoder one). This fact lead that we cannot use BERT \textit{as-it-is} for text generation. Nevertheless, there are some efforts to create new BERT-based algorithms for this purpose (the main idea consists of adding a decoder-transformer in the end). For example, the recent paper CG-BERT \cite{xia2020cgbert} looks very promising in this task, but we could not include it in this article due to lack of time and resources to train the model. The main idea of CG-BERT is to use half layers of BERT as encoders and half layers as decoders.
	
	In addition to BERT and GPT-2, we tried to implement a simple Transformer model to be capable of generating a conversation. The result is poor since the training data is really small as well as the resources to train the model. You can read the details in the \hyperref[sec:appendixTransformer]{appendix}.
	
	\section{Models comparison}
	
	\subsection{Architecture and pre-training}
	
	One difference that we encounter between BERT and GPT-2 is in the architecture: despite both are transformers-based architectures, it uses encoders in the case of BERT and, in the case of GPT-2, decoders. This particularity makes BERT and GPT-2 to understand text information in a different way and, in consecuence, their performance variate depending on the dataset.
	
	We find another difference in the pre-training method: GPT-2 aims to predict the next word in a sentence, while BERT is trained in NSP and MLM. We should keep in mind these pre-training methods when we use transfer learning to fine-tune a model.
	
	\subsection{Performance}
	
	In order to evaluate BERT and GPT-2, authors prepared different versions of both models (with different number of parameters) and fine-tuned them in different downstream tasks. One version is BERT Base, which has the same number of parameters as GPT (the predecesor of GPT-2) for comparison purposes. Unfortunately, we could not find common benchmarks between BERT and GPT-2 in the same datasets, so we only can compare BERT and GPT in GLUE and SWAG tasks.
	
	GLUE \cite{wang2018glue}, which consists of a wide list of NLP tasks (sentiment analysis, similarity, NLI, question-answering...), is one of the dataset where BERT was fine-tuned. In this dataset, BERT Base already outperformed GPT (and prior approaches such as BiLSTM+ELMo) and BERT Large got an even better result\footnote{GLUE leaderboard: \url{https://gluebenchmark.com/leaderboard}}, thus it achieved the status SOTA in this collection of tasks.
	
	SWAG \cite{zellers2018swagaf} consists of 113k multiple choice questions about grounded situations: given a question/situation, we have to pick the correct answer among four choices. Identically as GLUE, BERT outperformed GPT (BERT Large improved GPT score by $8.3\%$). It is also surprising that BERT Large got a better score than a human expert in a small subset of $100$ questions.
	
	\section{Related models}
	
	The models described in this article, BERT and GPT-2, demonstrate the benefits of large scale language modeling. Both papers leverage advances in compute and available text corpora to significantly surpass state-of-the-art performance in Natural Language Understanding (NLU), modeling, and generation. Currently, there are new models (based on BERT and GPT-2 architectures) which get even better results. Some of them are RoBERTa (robustly optimized version of BERT with larger training data) \cite{liu2019roberta}, ALBERT (lite version of BERT) \cite{lan2019albert}, DistilBERT (also a reduced version of BERT) \cite{sanh2019distilbert} and StructBERT (incorporate language structures into BERT pre-training) \cite{wang2019structbert} \footnote{In SQuAD, ALBERT has the SOTA status: \url{https://rajpurkar.github.io/SQuAD-explorer/}.} \footnote{In GLUE, StructBERT achieves the second best score: \url{https://gluebenchmark.com/leaderboard}} \footnote{In SWAG, RoBERTa is in the second position in the leaderboard: \url{https://leaderboard.allenai.org/swag/submissions/public}}.
	
	Appart from the mentioned BERT-based models, Nvidia published Megatron-LM \cite{shoeybi2019megatronlm} in 2019. In a few words, it is a simple and efficient model parallel approach by modificating existing PyTorch transformer implementations. They released a $3.9$ billion parameters version of BERT and $8.3$ of GPT-2. Both Megatron models got better scores than both vanilla BERT and GPT-2 models in the same tasks.
	
	\begin{figure}[ht]
		\centering
		\includegraphics[width=1.0\textwidth]{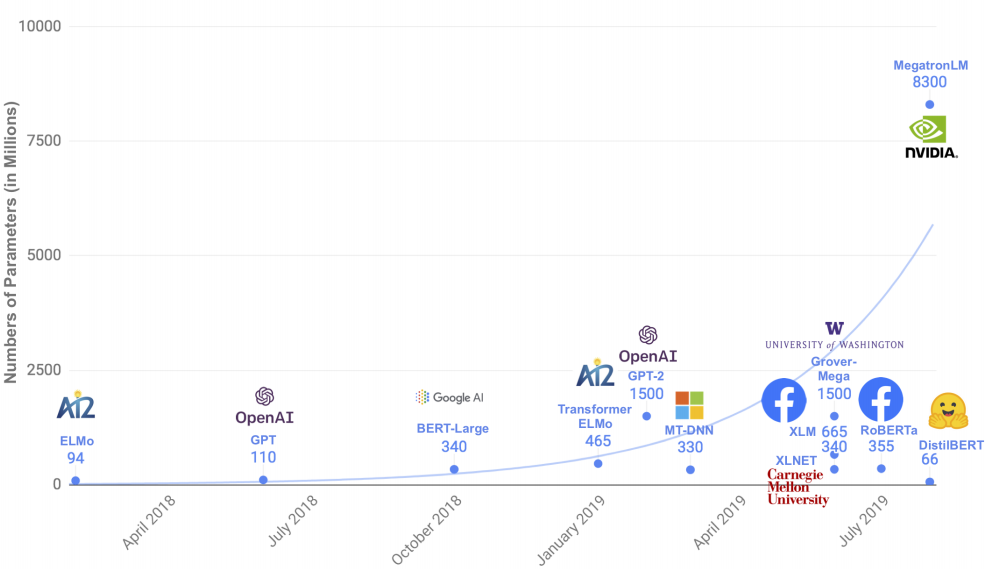}
		\caption{Evolution of number of parameters described by authors of DistilBERT \cite{sanh2019distilbert}.}
		\label{fig:NLPParamsEvolution}
	\end{figure}
	
	Additionally, OpenAI published recently a new model called GPT-3 (it uses the same architecture as GPT-2) \cite{brown2020language}. The largest version of this novel model has $175$ billion parameters which is 10 times more than any previous non-sparse language mode. However, despite the strong quantitative and qualitative improvements of GPT-3, particularly compared to its direct predecessor GPT-2, it still has notable weaknesses in text synthesis and several NLP tasks (they are extensively described in the paper). Unfortunately, we could not include text samples generated by GPT-3 in this article due to OpenAI has not released the model yet.
	
	\section{Conclusions}
	
	Transformers disrupted sequence-based deep learning significantly. The two variants of transformer models that we showed, BERT and GPT-2, outperform previous approaches with RNNs. In both cases, models take advantage of Attention layers, which selectively weight different elements in input data, and can achieve state-of-the-art performance on many language modelling benchmarks.
	
	Although there are some situations where the output of Transformers-based models is close to the human quality, we find that the output quality depends on how familiar is the model with the topic. For instance, GPT-2 is not able to generate a high-quality text when we provide an uncommon context (such as the recent COVID-19 pandemic or a new popular person). In the case of BERT, we observe that the model tend to answer incorrectly (in question-answering tasks) when we use homonyms or synonyms instead of the same words as the context. These experiments show us that still computers are distant from fully understanding of unstructured texts written in human languages.
	
	We observe a trend to build larger models in the NLP field, so they can capture more information from unstructured texts. Furthermore, unlikely humans, they need to be trained with tons of data. In terms of zero-shot or one-shot tasks, we need to improve the efficiency of NLP algorithms since they see much more text than a human sees in their lifetime.
	
	In general, we observe that despite these models achieve better performance than us in specific tasks, the common sense of humans produces better results rather than any deep learning model. Also, we cannot ignore the fact that we learn from our daily life, but it is difficult to measure what type of data should we provide to a model (during pre-training) to get a similar knowledge. This is the most likely reason deep learning models fail to produce coherent texts.
	
	In addition to the previous points, we observe a growing need to pre-train a model per language (BERT, GPT-2 or any other model). Apart from the grammar rules of each language (or, for example, the evident difference of RTL and LTR languages), there is knowledge which is intrinsic to the language: polite expressions, informal style... If we want to obtain human-quality in text generation, we will need large dataset not only in English but also in other languages (we cannot use translated data).
	
	In summary, we will see more transformer-based models in the future. They have demonstrated superiority in parallel computation and modelling long-range dependencies, compared to RNNs (such as LSTM). Still, we do not know which is the best approach to pre-train them as well as how to reproduce the human common sense (this is mandatory in text generation). Thus it is very likely that we are going to see not only new pre-training methods but also larger models.

	%----------------------------------------------------------------------------------------
	%	BIBLIOGRAPHY
	%----------------------------------------------------------------------------------------
	
	\newpage
	\renewcommand{\refname}{\spacedlowsmallcaps{References}} % For modifying the bibliography heading
	\addcontentsline{toc}{section}{References}

	\bibliography{references} 
	\bibliographystyle{unsrt}

	%----------------------------------------------------------------------------------------
	%	APPENDIX
	%----------------------------------------------------------------------------------------
	
	\newpage
	
	\appendix
	\addcontentsline{toc}{section}{Appendix}
	\section{Appendix: Experiments results}
	\label{sec:appendixExperiments}
	
	\subsection{Infer masked token}
	
	Despite some options could fit into philosophical or rare situations, we believe that they are uncommon options to replace the masked token. It also occurs that some options are grammatically correct, but they do not fit in the context of the sentence. In both cases, we note them as ``incorrect''.\newline
	
	\textbf{English}.
	
	\def\arraystretch{1.5}
	\begin{center}
		\begin{tabular}{||m{0.35\textwidth}|m{0.35\textwidth}|m{0.18\textwidth}||} 
			\hline
			\textbf{Sentence} & \textbf{Options} & \textbf{Notes} \\
			\hline\hline
			Hello, I'm a [MASK] model. & \begin{itemize}[noitemsep,parsep=0pt,partopsep=0pt]
				\item model $6.66\%$
				\item real $4.59\%$
				\item business $3.30\%$
				\item mathematical $3.16\%$
				\item new $2.78\%$
			\end{itemize} & All correct. \\
			\hline
			Today is [MASK]. & \begin{itemize}[noitemsep,parsep=0pt,partopsep=0pt]
				\item demolished $19.43\%$
				\item unknown $8.44\%$
				\item closed $5.93\%$
				\item abandoned $2.82\%$
				\item active $2.72\%$
			\end{itemize} & Only the third option, \textit{closed}, seems correct. \\
			\hline
			[MASK] is a good idea. & \begin{itemize}[noitemsep,parsep=0pt,partopsep=0pt]
				\item It $48.57\%$
				\item This $30.64\%$
				\item That $7.16\%$
				\item There $2.25\%$
				\item it $0.70\%$
			\end{itemize} & All correct. \\
			\hline
			The doctor ran to the emergency room to see [MASK] patient. & \begin{itemize}[noitemsep,parsep=0pt,partopsep=0pt]
				\item the $88.97\%$
				\item a $4.84\%$
				\item another $2.83\%$
				\item his $1.07\%$
				\item her $0.76\%$
			\end{itemize} & All correct. \\
			\hline
			The doctor ran to the emergency room to see the [MASK]. & \begin{itemize}[noitemsep,parsep=0pt,partopsep=0pt]
				\item incident $4.33\%$
				\item accident $3.85\%$
				\item explosion $3.77\%$
				\item scene $2.64\%$
				\item crash $2.49\%$
			\end{itemize} & All correct. \\
			\hline
		\end{tabular}
	\end{center}
	
	\newpage
	\textbf{Russian}.
	
	\def\arraystretch{1.5}
	\begin{center}
		\begin{tabular}{||m{0.35\textwidth}|m{0.35\textwidth}|m{0.18\textwidth}||} 
			\hline
			\textbf{Sentence} & \textbf{Options} & \textbf{Notes} \\
			\hline\hline
			\foreignlanguage{russian}{Я считаю, что Настя очень [MASK] человек.} & \begin{itemize}[noitemsep,parsep=0pt,partopsep=0pt]
				\item \foreignlanguage{russian}{молодой} $54.84\%$
				\item \foreignlanguage{russian}{большой} $20.15\%$
				\item \foreignlanguage{russian}{великий} $5.03\%$
				\item \foreignlanguage{russian}{лучший} $4.91\%$
				\item \foreignlanguage{russian}{новый} $1.64\%$
			\end{itemize} & All correct except the last option ( \foreignlanguage{russian}{новый}). \\
			\hline
			\foreignlanguage{russian}{Я очень люблю [MASK]. Я рисую днями и ночами.} & \begin{itemize}[noitemsep,parsep=0pt,partopsep=0pt]
				\item \foreignlanguage{russian}{музыку} $13.82\%$
				\item \foreignlanguage{russian}{книги} $3.11\%$
				\item \foreignlanguage{russian}{жизнь} $2.64\%$
				\item \foreignlanguage{russian}{людей} $2.58\%$
				\item \foreignlanguage{russian}{игру} $2.34\%$
			\end{itemize} & All incorrect. \\
			\hline
			\foreignlanguage{russian}{Он обладает многими [MASK].} & \begin{itemize}[noitemsep,parsep=0pt,partopsep=0pt]
				\item \foreignlanguage{russian}{медалями} $44.46\%$
				\item \foreignlanguage{russian}{орденами} $18.62\%$
				\item \foreignlanguage{russian}{людьми} $12.01\%$
				\item \foreignlanguage{russian}{словами} $5.61\%$
				\item \foreignlanguage{russian}{результатами} $3.22\%$
			\end{itemize} & Only the first (\foreignlanguage{russian}{медалями}) and the second (\foreignlanguage{russian}{орденами}) options are correct.\\
			\hline
		\end{tabular}
	\end{center}
	
	\textbf{Spanish}.
	
	\def\arraystretch{1.5}
	\begin{center}
		\begin{tabular}{||m{0.35\textwidth}|m{0.35\textwidth}|m{0.18\textwidth}||} 
			\hline
			\textbf{Sentence} & \textbf{Options} & \textbf{Notes} \\
			\hline\hline
			Ayer estuve paseando y me encontré con [MASK]. & \begin{itemize}[noitemsep,parsep=0pt,partopsep=0pt]
				\item \#migo $34.68\%$ (\textit{conmigo})
				\item él $14.68\%$
				\item ella $7.43\%$
				\item ellos $3.16\%$
				\item Dios $2.60\%$
			\end{itemize} & All correct except the first option (\textit{conmigo}).\\
			\hline
			Si tuviera [MASK] podría irme de vacaciones. & \begin{itemize}[noitemsep,parsep=0pt,partopsep=0pt]
				\item dinero $14.99\%$
				\item , $6.45\%$
				\item tiempo $5.54\%$
				\item años $4.50\%$
				\item problemas $2.79\%$
			\end{itemize} & The first three options are correct. \\
			\hline
			Necesito [MASK] para aprobar el examen. & \begin{itemize}[noitemsep,parsep=0pt,partopsep=0pt]
				\item tiempo $39.88\%$
				\item dinero $4.38\%$
				\item pruebas $3.30\%$
				\item información $3.10\%$
				\item recursos $2.57\%$
			\end{itemize} & The first (\textit{tiempo}) and the last two options (\textit{información} and \textit{recursos}) are correct. \\
			\hline
			Voy a ir con mis amigos al [MASK]. & \begin{itemize}[noitemsep,parsep=0pt,partopsep=0pt]
				\item colegio $7.99\%$
				\item pueblo $5.74\%$
				\item mar $3.19\%$
				\item mundo $2.57\%$
				\item hotel $2.16\%$
			\end{itemize} & All correct except the forth option (\textit{mundo}). \\
			\hline
		\end{tabular}
	\end{center}
	
	\subsection{Question-answering}
	
	\textbf{English}. The context is the Napoleon's biography extracted from Wikipedia\footnote{\url{https://en.wikipedia.org/wiki/Napoleon}}.
	
	\begin{center}
		\noindent\fbox{%
			\parbox{0.98\textwidth}{%
				Napoleon Bonaparte (15 August 1769 - 5 May 1821) was a French statesman and military leader who became notorious as an artillery commander during the French Revolution. He led many successful campaigns during the French Revolutionary Wars and was Emperor of the French as Napoleon I from 1804 until 1814 and again briefly in 1815 during the Hundred Days. Napoleon dominated European and global affairs for more than a decade while leading France against a series of coalitions during the Napoleonic Wars. He won many of these wars and a vast majority of his battles, building a large empire that ruled over much of continental Europe before its final collapse in 1815. He is regarded as one of the greatest military commanders in history, and his wars and campaigns are studied at military schools worldwide. Napoleon's political and cultural legacy has made him one of the most celebrated and controversial leaders in human history.
			}%
		}
	\end{center}
	
	\def\arraystretch{1.5}
	\begin{center}
		\begin{tabular}{||m{0.35\textwidth}|m{0.35\textwidth}|m{0.18\textwidth}||} 
			\hline
			\textbf{Question} & \textbf{Answer} & \textbf{Notes} \\
			\hline\hline
			When did Napoleon become Emperor? & 1804 until 1814 & Correct, but it should answer only \textit{1804}. \\
			\hline
			Who was Napoleon Bonaparte? & Napoleon Bonaparte (15 August 1769 - 5 May 1821), born Napoleone di Buonaparte & Incorrect. \\
			\hline
			Where are studied his wars and campaigns? & military schools worldwide & Correct. \\
			\hline
			What is studied about him? & military schools worldwide & Incorrect. \\
			\hline
			How long Napoleon dominated Europe? & for more than a decade & Correct. \\
			\hline
		\end{tabular}
	\end{center}
	
	\textbf{Russian}. Context extracted from Wikipedia about Higher School of Economics\footnote{\url{https://ru.wikipedia.org/wiki/\%D0\%92\%D1\%8B\%D1\%81\%D1\%88\%D0\%B0\%D1\%8F_\%D1\%88\%D0\%BA\%D0\%BE\%D0\%BB\%D0\%B0_\%D1\%8D\%D0\%BA\%D0\%BE\%D0\%BD\%D0\%BE\%D0\%BC\%D0\%B8\%D0\%BA\%D0\%B8}}.
	
	\begin{center}
		\noindent\fbox{%
			\parbox{0.98\textwidth}{%
				\foreignlanguage{russian}{Национальный исследовательский университет «Высшая школа экономики» (НИУ ВШЭ; Вышка) — автономное учреждение, федеральное государственное высшее учебное заведение. ВШЭ создана в 1992 году, нынешний статус носит с 2009 года. Основной кампус находится в Москве, ещё три — в Нижнем Новгороде, Санкт-Петербурге и Перми.}
			}%
		}
	\end{center}
	
	\def\arraystretch{1.5}
	\begin{center}
		\begin{tabular}{||m{0.35\textwidth}|m{0.35\textwidth}|m{0.18\textwidth}||} 
			\hline
			\textbf{Question} & \textbf{Answer} & \textbf{Notes} \\
			\hline\hline
			\foreignlanguage{russian}{Когда была основана Вышка?} (\textit{When was HSE founded?}) & \foreignlanguage{russian}{1992 году} (\textit{1992 year}) & Correct answer, but it is gramatically incorrect. \\
			\hline
			\foreignlanguage{russian}{Где расположены кампусы?} (\textit{Where are the campuses located?}) & \foreignlanguage{russian}{Москве} (\textit{Moscow}) & Partial answer. \\
			\hline
			\foreignlanguage{russian}{Что такое Высшая школа экономики?} (\textit{What is the Higher School of Economics?}) & \foreignlanguage{russian}{Национальный исследовательский университет} (\textit{National Research University}) & Correct, but it is not accurate. \\
			\hline
			\foreignlanguage{russian}{Что такое аббревиатура Высшей школы экономики?} (\textit{What is the acronym for the Higher School of Economics?}) & \foreignlanguage{russian}{Национальный исследовательский университет «Высшая школа экономики»} (\textit{National Research University Higher School of Economics}) & Incorrect. \\
			\hline
		\end{tabular}
	\end{center}
	
	\textbf{Spanish}. Description about Sagrada Familia extracted from Wikipedia\footnote{\url{https://es.wikipedia.org/wiki/Templo_Expiatorio_de_la_Sagrada_Familia}}.
	
	\begin{center}
		\noindent\fbox{%
			\parbox{0.98\textwidth}{%
				El Tempo Expiatorio de la Sagrada Familia (en catalán, Temple Expiatori de la Sagrada Família), conocido simplemente como la Sagrada Familia, es una basílica católica de Barcelona (España), diseñada por el arquitecto Antoni Gaudí. Iniciada en 1882, todavía está en construcción. Es la obra maestra de Gaudí, y el máximo exponente de la arquitectura modernista catalana. Es uno de los monumentos más visitados de España, junto al Museo del Prado y la Alhambra de Granada, y es la iglesia más visitada de Europa tras la basílica de San Pedro del Vaticano. Cuando esté finalizada será la iglesia cristiana más alta del mundo.
			}%
		}
	\end{center}
	
	\def\arraystretch{1.5}
	\begin{center}
		\begin{tabular}{||m{0.35\textwidth}|m{0.35\textwidth}|m{0.18\textwidth}||} 
			\hline
			\textbf{Question} & \textbf{Answer} & \textbf{Notes} \\
			\hline\hline
			¿Cómo se escribe en catalán? (\textit{How is it written in Catalan?}) & Temple Expiatori de la Sagrada Família & Correct \\
			\hline
			¿Quién diseñó la iglesia? (\textit{Who designed the church?}) & Antoni Gaudí & Correct \\
			\hline
			¿Cuándo comenzó la construcción de la Sagrada Familia? (\textit{When did the construction of Sagrada Familia began?}) & 1882 & Correct \\
			\hline
			¿Qué lugar cristiano recibe más visitas que la Sagrada Familia? (\textit{Which christian place receives more visits than Sagrada Familia?}) & \textit{It outputs the whole context.} & Incorrect \\
			\hline
			¿Qué otros monumentos son muy visitados por turistas en España? (\textit{Which other monuments are highly visited by tourists in Spain?}) & Museo del Prado y la Alhambra de Granada & Correct \\
			\hline
		\end{tabular}
	\end{center}
	
	\subsection{Conditional text generation}
	
	The following synthetic texts were generated by GPT-2. We fed the model with an initial text and then we let GPT-2 to finish it. These text samples are written only in English due GPT-2 end up in nonsense outputs using other languages. Note that the model does not give the same output everytime we run it, so you might obtain a different output.
	
	\begin{center}
		\noindent\fbox{%
			\parbox{0.98\textwidth}{%
				\textbf{Last week, many people reported that they saw a unicorn at the park.} It turned out to be a real unicorn, but it was just a fake.  The unicorn was made of foam and plastic. The foam was made of polystyrene, which is a plastic that is commonly used in construction materials. The fake unicorn was made of foam, which has been known to be a very effective material for creating fake animals, such as this one. The foam was made of polystyrene, [...]
			}%
		}
	\end{center}
	
	\begin{center}
		\noindent\fbox{%
			\parbox{0.98\textwidth}{%
				\textbf{In the last months, a new virus spread worldwide and changed our daily lifes. This new virus, called} Ebola, appeared in three countries, and has killed more than 3,000 people in the four countries: Guinea, Liberia and Sierra Leone. The virus is spread through direct contact with the fluids of infected persons, which can include body fluids, such as blood, sweat and tears…  It is also spread through contaminated food and water. When Ebola arrives in a country, a lot of people [...]
			}%
		}
	\end{center}
	
	\begin{center}
		\noindent\fbox{%
			\parbox{0.98\textwidth}{%
				\textbf{Q: How would you describe to Maria?}
				
				\textbf{A:} I'd say… I'd say that she 's like very strong and very resilient.
				
				Q: Thank you very much.
				
				A: Oh, thanks for the offer to talk.
			}%
		}
	\end{center}
	
	\begin{center}
		\noindent\fbox{%
			\parbox{0.98\textwidth}{%
				\textbf{Q: What do you think about the death penalty?}
				
				\textbf{A:} I am not an advocate for the death penalty. I am not someone who believes that this life should be taken and then taken over again, which of course is the case with every case when we see a murder. It is my belief that a fair penalty should be used, and that if the evidence was not sufficient to prove that the person committed the actual crime, then there should be a life sentence.
			}%
		}
	\end{center}
	
	\begin{center}
		\noindent\fbox{%
			\parbox{0.98\textwidth}{%
				\textbf{Teacher: Good morning everyone.}
				
				\textbf{Students: Good morning teacher.}
				
				\textbf{Teacher:} It's my pleasure to welcome you all here today.
				
				Students: Yeah.
				
				Teacher: I want to talk about some things that you all do.
				
				Students: Yeah.
				
				Teacher: You guys always seem to have such great enthusiasm.
				
				Students: Yeah.
				
				Teacher: So, we'll start with our first topic. What's your favorite type of book to read?
			}%
		}
	\end{center}
	
	\newpage
	
	\section{Appendix: Simple Transformer model}
	\label{sec:appendixTransformer}
	
\begin{figure}[ht]
		\centering
		\includegraphics[width=0.7\textwidth]{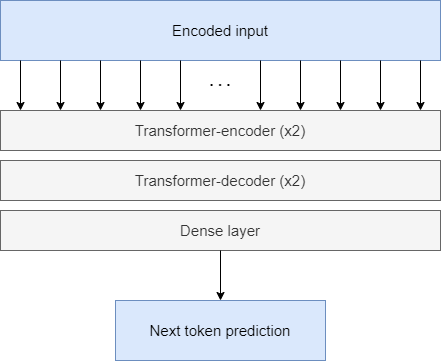}
		\caption{Visual description of our Transformer model.}
		\label{fig:ExperimentTransformersModel}
	\end{figure}

	We implemented a simple model (see Figure \ref{fig:ExperimentTransformersModel}) to figure out whether we are able of generating text based on training data. For this purpose, we took an English dialogue written by Shakespeare\footnote{We took the dialogue from this TensorFlow tutorial: \url{https://www.tensorflow.org/tutorials/text/text_generation}}.
	
	The model consists of 2 transformer-encoders and 2 transformer-decoders to capture the text information, and a fully-connected layer to obtain the prediction for the next token in the text. We kept the model simple due to two reasons:
	
	\begin{itemize}
		\item We do not have enough resources to train a large model like BERT or GPT-2.
		\item If the model was larger, we would also need a bigger dataset to train all its parameters.
	\end{itemize}
	
	Unfortunately, the model predictions are really bad in the sense that it outputs nonsense tokens to continue a text. Here you can see the sample text that the model generated from the initial text ``\textit{ROMEO:}''. It is very likely  that we need to increase the model size to better capture the text information (in consecuence, we need to use more data to train it) and, therefore, generate a coherent text.
	
	\begin{center}
		\noindent\fbox{%
			\parbox{0.98\textwidth}{%
				\textbf{ROMEO:}  took perform still sets big,KATHARINA'd me my heart
				
				l belie to harpets'd upon me
				
				And sure darkness under'd statutes.
				
				deceive them or orderly and likely body will f roodFor
				
				Masteredon'd conclude else aged asTherefore
				
				The suffer hiswittedWould being royalties and to ouroan as which and than him
				
				The goddess wasYour things curst hag that they admirings that how of they
				
				twere to spirit:
				
				linger OF glory of follow
				
				ised greatestar
			}%
		}
	\end{center}

	%----------------------------------------------------------------------------------------
	
\end{document}

%% file: structure.tex
\usepackage[
nochapters, % Turn off chapters since this is an article        
beramono, % Use the Bera Mono font for monospaced text (\texttt)
eulermath,% Use the Euler font for mathematics
pdfspacing, % Makes use of pdftex’ letter spacing capabilities via the microtype package
dottedtoc % Dotted lines leading to the page numbers in the table of contents
]{classicthesis} % The layout is based on the Classic Thesis style

\usepackage{indentfirst}

\usepackage{arsclassica} % Modifies the Classic Thesis package

\usepackage[T2A,T1]{fontenc} % Use 8-bit encoding that has 256 glyphs

\usepackage[utf8]{inputenc} % Required for including letters with accents
\usepackage[russian,spanish,english]{babel}

\usepackage{graphicx} % Required for including images
\graphicspath{{Figures/}} % Set the default folder for images

\usepackage{enumitem} % Required for manipulating the whitespace between and within lists

\usepackage{subfig} % Required for creating figures with multiple parts (subfigures)

\usepackage{amsmath,amssymb,amsthm} % For including math equations, theorems, symbols, etc

\usepackage{varioref} % More descriptive referencing

\usepackage{hyperref}
\makeatletter
\def\ttl@useclass#1#2{%
  \@ifstar
    {\ttl@labelfalse\@dblarg{#1{#2}}}
    {\ttl@labeltrue\@dblarg{#1{#2}}}}
\makeatother

\usepackage{booktabs}

\usepackage{array}

\usepackage[none]{hyphenat} % No hyphenation

\makeatletter
\def\supervisor#1{\gdef\@supervisor{#1}}
\def\@maketitle{
  \newpage
  \null
  \vskip 2em
  \begin{center}
  \let \footnote \thanks
    {\large
    	\@title \\
    	\vskip 1.0em
    	\@subtitle \par}
    \vskip 1.5em
    {\large
      \lineskip .5em
      \begin{tabular}[t]{c}
        \@author
      \end{tabular}\par}
    \vskip 1em
    {\large \@date}
  \end{center}
  \par
  \vskip 1.5em}
\makeatother

\theoremstyle{definition} % Define theorem styles here based on the definition style (used for definitions and examples)

\theoremstyle{plain} % Define theorem styles here based on the plain style (used for theorems, lemmas, propositions)

\theoremstyle{remark} % Define theorem styles here based on the remark style (used for remarks and notes)

\hypersetup{
colorlinks=true, breaklinks=true, bookmarks=true,bookmarksnumbered,
urlcolor=webbrown, linkcolor=RoyalBlue, citecolor=webgreen, % Link colors
pdftitle={}, % PDF title
pdfauthor={\textcopyright}, % PDF Author
pdfsubject={}, % PDF Subject
pdfkeywords={}, % PDF Keywords
pdfcreator={pdfLaTeX}, % PDF Creator
pdfproducer={LaTeX with hyperref and ClassicThesis} % PDF producer
}